\documentclass[10pt,twocolumn,letterpaper]{article}

\usepackage{preprint}

\usepackage{multirow}
\usepackage{colortbl}
\usepackage{algorithm}
\usepackage{algpseudocode}
\makeatletter
\renewcommand\fs@ruled{\def\@fs@cfont{\bfseries}\let\@fs@capt\floatc@ruled
  \def\@fs@pre{{\normalcolor\hrule height.8pt depth0pt}\kern2pt}%
  \def\@fs@post{\kern2pt{\normalcolor\hrule}\relax}%
  \def\@fs@mid{\kern2pt{\normalcolor\hrule}\kern2pt}%
  \let\@fs@iftopcapt\iftrue}
\floatstyle{ruled}\restylefloat{algorithm}
\makeatother
\DeclareMathOperator{\tr}{tr}
\newcommand{\method}{LAQ\xspace}

\newcommand{\bOmega}{\boldsymbol{\Omega}}
\newcommand{\bone}{\mathbf{1}}
\newcommand{\be}{\mathbf{e}}
\newcommand{\bz}{\mathbf{z}}
\newcommand{\ba}{\mathbf{a}}
\newcommand{\bo}{\mathbf{o}}
\newcommand{\bx}{\mathbf{x}}
\newcommand{\bW}{\mathbf{W}}
\newcommand{\bH}{\mathbf{H}}
\newcommand{\bM}{\mathbf{M}}
\newcommand{\bJ}{\mathbf{J}}
\newcommand{\bD}{\mathbf{D}}

\definecolor{linkblue}{rgb}{0.21,0.49,0.74}
\usepackage[pagebackref,breaklinks,colorlinks,allcolors=linkblue]{hyperref}

\title{Don't Lose the Future: Latent-Aware 4-Bit Quantization\\of World Action Models}

\author{Yichen Guo$^{1}$\quad Yanlong Chen$^{1}$\quad Xinyue Zhang$^{1}$\quad Mohan Liu$^{1}$\quad Song Zhang$^{2}$\quad Yawei Li$^{1,*}$\\[0.4em]
{\normalsize $^{1}$Nanyang Technological University\qquad $^{2}$Nanjing University}\\[0.2em]
{\small $^{*}$Corresponding author}
}

\begin{document}
\maketitle
\begin{abstract}
World action models (WAMs) predict how a scene will evolve and condition their actions on this predicted future.
Deploying them on a robot's onboard computer calls for 4-bit weights, activations and key--value caches (W4A4KV4).
Existing 4-bit quantizers for WAMs preserve the generated action, and estimate sensitivity one denoising step at a time.
However, they fail to preserve the fidelity of the predicted future: once the action pathway is protected, the remaining action error enters mostly through a degraded future that none of them deliberately protects.
We propose \method, a latent-aware quantization framework that preserves the predicted future where it is written, stored and read.
Specifically, we introduce a reader metric, which measures future error by its effect on the action, and a read count, which weights each rounding error by the number of denoising steps that read it.
Together they give the allocation of precision over activations and caches a closed-form solution.
We further develop gaze-guided refinement, which gives the tokens the action attends to a second 4-bit pass in the denoising loop and in the cache.
All weights stay at 4 bits, with no high-precision branch and no per-task recalibration.
On the LIBERO-Plus benchmark, \method reaches 76.4\% and 83.3\% average success on Fast-WAM and Cosmos-Policy, within 1.2 percentage points of the 16-bit models and all above the baselines with 16-bit KV caches.
It compresses the model by 3.8$\times$, reduces the peak GPU memory of the whole policy by 2.3$\times$, and runs 1.66$\times$ faster than the 16-bit model. On a real AgileX PiPER arm, it reaches 47.0\% average success over three tasks, against 52.0\% for the 16-bit model and 30.0\%  higher than QuaRot.
\end{abstract}

\section{Introduction}
\label{sec:intro}

\begin{figure}[t]
  \centering
  \includegraphics[width=\linewidth]{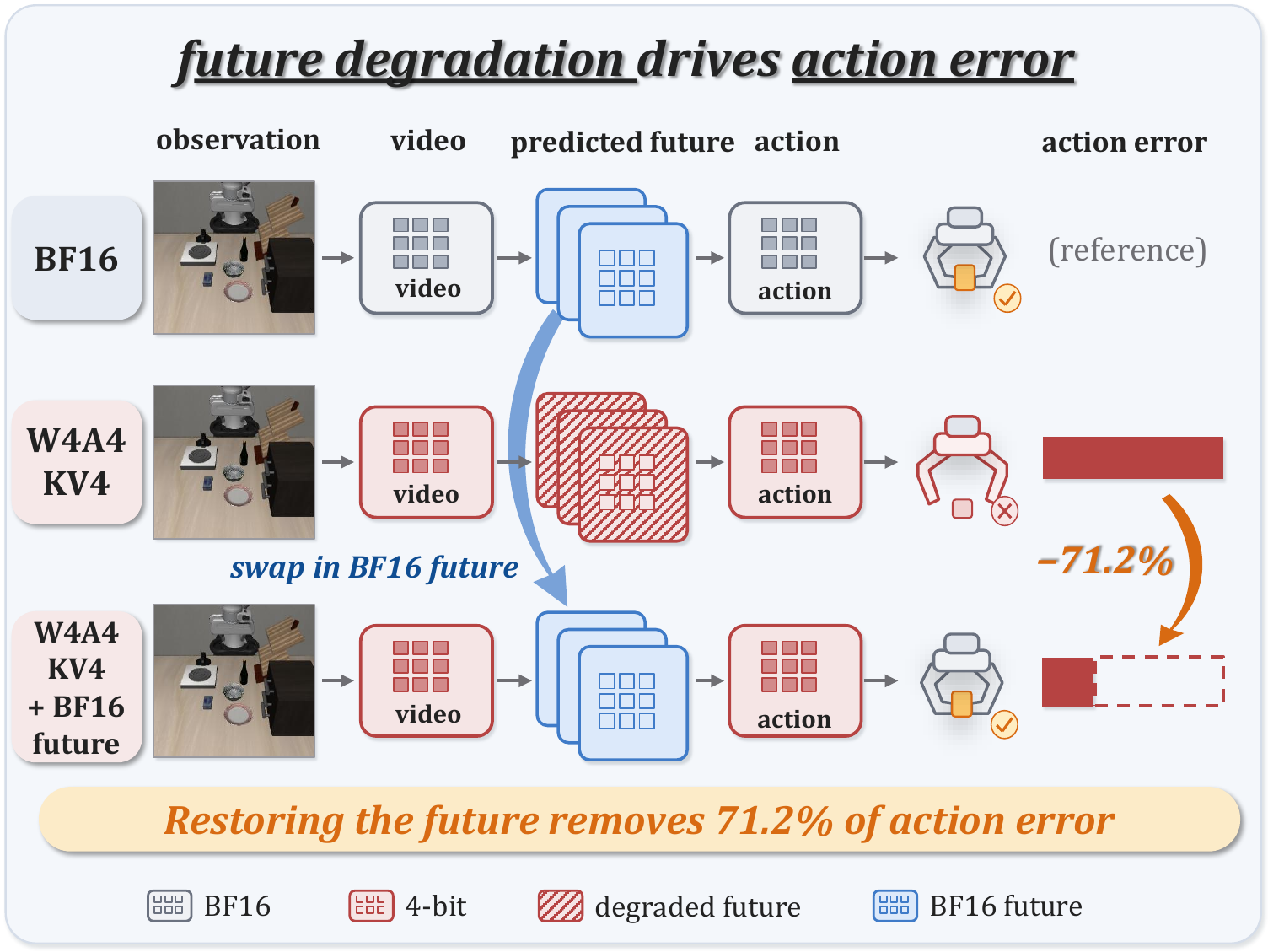}
  \caption{Protecting the action is not enough.
  In a 4-bit WAM whose action pathway is kept at higher precision, the predicted future still degrades, and 71.2\% of the remaining action error enters through it (Fast-WAM, W4A4KV4).
  \method preserves the future where it is written, stored and read.}
  \label{fig:teaser}
\end{figure}

World action models (WAMs) jointly predict how a scene will evolve and how to act~\cite{yuan2026fastwam,kim2026cosmospolicy,li2026lingbotva,ye2026dreamzero}.
Trained together with a video generator, they condition their actions on a predicted future and inherit the visual priors of large video backbones, and also their size.
Fast-WAM~\cite{yuan2026fastwam} has 6B parameters, and its BF16 weights alone occupy 12\,GB of the 16\,GB that an onboard computer such as the NVIDIA Jetson Orin NX shares with the rest of the robot stack.
Post-training quantization removes most of this cost without retraining: a fully 4-bit pipeline stores the weights, the activations and the key--value caches of attention at 4 bits (W4A4KV4)~\cite{ashkboos2024quarot,liu2025spinquant}.
These savings matter only if the quantized policy still acts reliably in closed loop.
Existing quantizers, whether built for language and diffusion models~\cite{xiao2023smoothquant,ashkboos2024quarot,liu2025spinquant,frantar2023gptq,zhao2025viditq,li2025svdquant} or recently for WAMs~\cite{zhou2026quantwams,akbari2026qwam,wang2026steerquant}, preserve each layer's output or the final action one denoising step at a time; none protects the predicted future on which the action depends, or the caches through which it is read at every denoising step.

We find that this omission is costly: existing quantizers fail to preserve the fidelity of the predicted future, and the loss reaches the action.
We call \emph{reader} the part of a WAM that maps the predicted future to the action, and ask what a 4-bit WAM still loses once its action pathway is protected.
Three observations on Fast-WAM answer it (\cref{sec:motivation}).
First, protecting the action is not enough: restoring the predicted future to full precision removes 71.2\% of the remaining action error (\cref{fig:teaser}).
Second, the future is damaged where it is written, and most by errors that are read many times: an error frozen in a weight, a cache or an unchanging token damages the future two to three times more than an error of the same energy that is redrawn at each denoising step.
Third, the reader depends on a narrow part of the future, stable in location but not in direction: the tokens it depends on agree across contexts while the directions do not.
A 4-bit WAM must therefore treat the predicted future, as the reader sees it, as a quantization target in its own right.

\begin{figure*}[t]
  \centering
  \includegraphics[width=\textwidth]{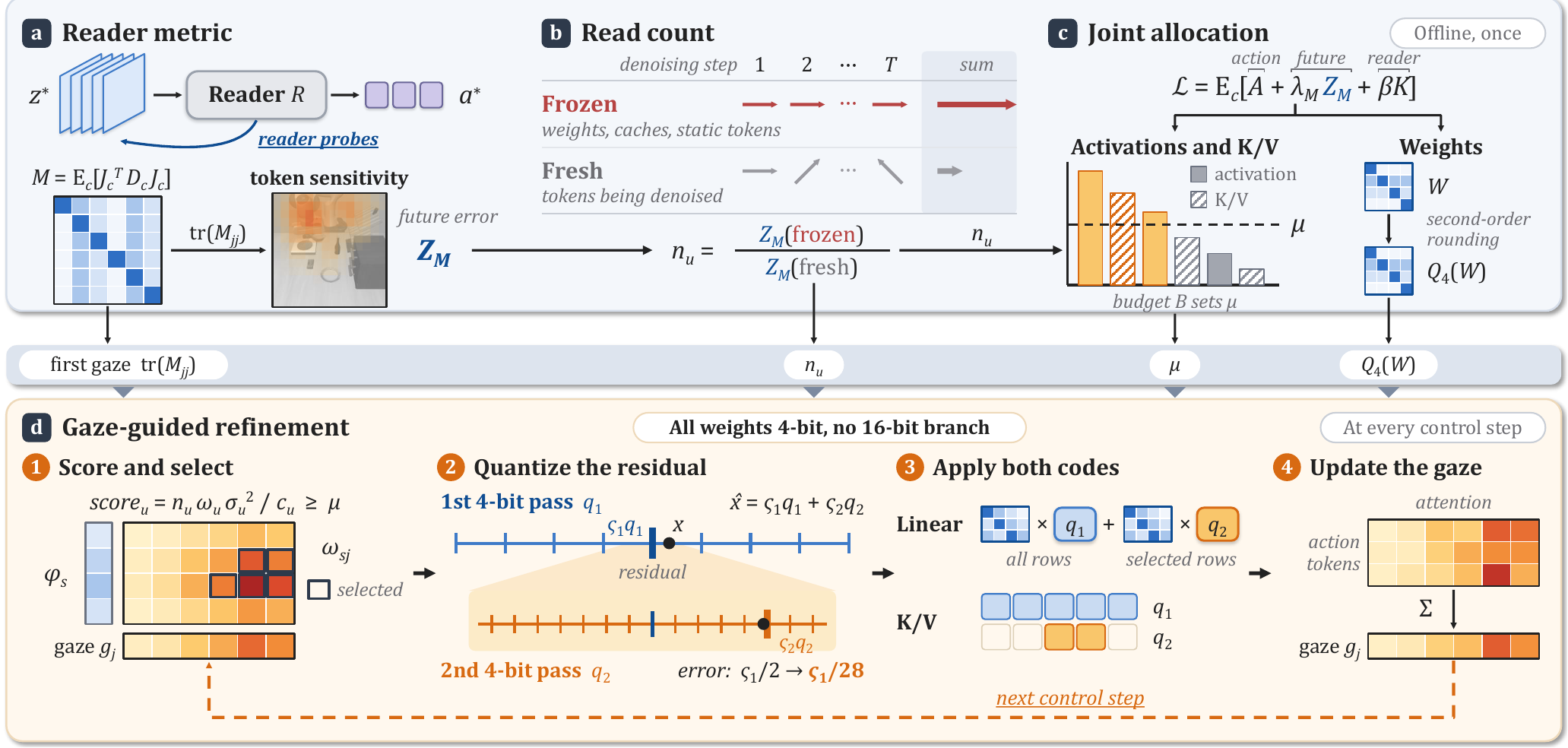}
  \caption{Overview of \method.
  Offline, \method estimates the reader metric $\bM$, which measures future error as the reader sees it~(a), and the read count $n_u$, the damage of a rounding error that is frozen across the denoising steps relative to one that is redrawn at every step~(b), and sets from one objective a threshold $\mu$ shared by activations and caches~(c).
  At every control step, gaze-guided refinement~(d) gives the tokens selected under the current gaze a second 4-bit code in the linear layers and in the cache, and the attention of the action tokens gives the gaze for the next step.}
  \label{fig:method}
\end{figure*}

Building on this diagnosis, we propose \method, a latent-aware quantization framework for W4A4KV4 world action models (\cref{fig:method}).
\method estimates a \emph{reader metric}, which measures future error by its effect on the action, and assigns every source of rounding error a \emph{read count}, the number of denoising steps over which one realization of the error is read, so that errors frozen in caches and unchanging tokens are weighted by their accumulated effect.
With both, the allocation of precision over activations and caches has a closed-form solution: a single threshold that decides which tokens deserve more precision in the denoising loop and in the cache.
\emph{Gaze-guided refinement} then gives those tokens a second 4-bit pass, and selects them at run time from the attention of the action tokens.
The quantized model stores every weight at 4 bits, has no high-precision branch, and needs no training and no recalibration for a new task.
In summary, our contributions are threefold:
\begin{itemize}
  \item We identify future-latent fidelity as a key quantization target for WAMs: once the action pathway is protected, the remaining action error of a 4-bit WAM enters through a degraded predicted future, which is damaged most by errors that are written once and read at every denoising step.
  \item We propose \method, a latent-aware quantization framework for W4A4KV4 WAMs. A reader metric and a read count give the allocation of precision over activations and caches a closed-form solution, and gaze-guided refinement protects the tokens the reader depends on with a second 4-bit pass.
  \item We evaluate \method on two WAMs in simulation and on a real AgileX PiPER arm. On LIBERO-Plus~\cite{fei2026liberoplus}, it reaches 76.4\% average success on Fast-WAM, within 1.2 points of BF16 and 3.4 points above the strongest baseline with 16-bit caches, and 83.3\% on Cosmos-Policy (BF16: 83.5\%), while compressing the model by 3.8$\times$. On three real-world tasks, it reaches an average success of 47.0\% (BF16: 52.0\%), compared with 17.0\% for QuaRot.
\end{itemize}

\section{Related Work}
\label{sec:related}

\noindent\textbf{World action models.}
World action models train a video generator jointly with an action model and condition actions on a predicted future~\cite{yuan2026fastwam,kim2026cosmospolicy,li2026lingbotva,ye2026dreamzero}, which extends vision-language-action policies~\cite{kim2024openvla,black2024pi0} with the visual priors, and the size, of video backbones.

\noindent\textbf{Post-training quantization.}
Second-order rounding compensates the error of each layer with the inverse Hessian of a layer-wise reconstruction objective~\cite{frantar2022obc,frantar2023gptq}, and activation-aware scaling~\cite{xiao2023smoothquant,lin2024awq} and rotations~\cite{ashkboos2024quarot,liu2025spinquant,chen2026prismquant} remove activation outliers, which brings language models to 4-bit weights, activations and key--value caches.
Mixed-precision methods allocate bit widths across layers~\cite{wang2019haq,malinovskii2025higgs}, and GuidedQuant weights the layer objective with gradients of the end loss~\cite{kim2025guidedquant}.
For diffusion and video generators, post-training methods correct errors that accumulate over the denoising loop~\cite{he2023ptqd}, handle timestep-dependent activations~\cite{zhao2025viditq} and absorb outliers into a 16-bit low-rank branch~\cite{li2025svdquant}; quantization-aware training reaches 4 bits and below for these generators~\cite{huang2026qvgen,li2026dsaquant} and for vision-language models~\cite{chen2026grace}.
Cache quantizers compress the key--value caches of video-language models~\cite{tao2025vidkv,tao2026windowquant} and of video generators and world models~\cite{xi2026qvg,zhao2026quantwm}, down to 2 bits and below.
These methods preserve layer outputs or generation quality; \method builds on rotation and second-order rounding, and changes what the quantizer preserves: the predicted future as the reader sees it.

\noindent\textbf{Quantization of robot policies and WAMs.}
QuantVLA and BitVLA compress vision-language-action policies with calibrated scales and 1-bit weights~\cite{zhang2026quantvla,wang2025bitvla}.
For WAMs, QuantWAMs~\cite{zhou2026quantwams}, Q-WAM~\cite{akbari2026qwam} and SteerQuant~\cite{wang2026steerquant} decide which channels, activation directions, token streams and denoising steps to protect by their impact on the action, and PreDE predicts the task degradation of a quantization configuration from offline action deviations~\cite{xu2026prede}.
They measure sensitivity to the final action or the training loss, one denoising step at a time.
\method instead targets the predicted future that the action is read from, counts how often each rounding error is read, and quantizes the caches.

\noindent\textbf{Token-level precision.}
Squat keeps the tokens with the highest attention at 8 bits~\cite{shen2024squat}, STaMP concentrates the activation energy in a few tokens that receive higher precision~\cite{federici2026stamp}, SemanticDialect re-quantizes the residual of salient tokens in video diffusion transformers~\cite{jang2026semanticdialect}, and ZipCache assigns cache precision by token saliency~\cite{he2024zipcache}.
In all of these token-level quantizers, a model's own attention or activation energy decides which tokens matter.
In a WAM, the predicted future is consumed by a different part of the model, and \method lets this reader decide: tokens are selected by their effect on the action and tracked at run time through the attention of the action tokens.

\section{Motivation}
\label{sec:motivation}

\subsection{Quantizing a WAM}
\label{sec:prelim}

\noindent\textbf{World action models.}
A WAM maps an observation $\bo$ and an instruction $\ell$ to a predicted future latent $\bz$ and an action chunk $\ba$, and the action is conditioned on $\bz$:
\begin{equation}
  \bz = G_{\bW}(\bo, \ell), \qquad \ba = R_{\bW}(\bo, \ell, \bz),
  \label{eq:wam}
\end{equation}
where $\bW$ collects all weights and both are produced by iterative denoising over $T$ steps.
We call $R$ the \emph{reader}, the computation that turns the predicted future into an action; it is well defined whether $\bz$ and $\ba$ are generated one after the other or denoised jointly (\cref{sec:exp:setup}).
We write $\bz^\star, \ba^\star$ for the outputs of the full-precision model and $\hat G, \hat R, \hat\bz, \hat\ba$ for those of a quantized one.

\noindent\textbf{4-bit quantization.}
A $b$-bit quantizer maps a tensor $X$ onto a scaled integer grid, $Q_b(X) = \varsigma\cdot\operatorname{round}(X/\varsigma)$, with one scale $\varsigma = \max|X| / (2^{b-1}-1)$ per group of entries.
Weights are quantized once offline and activations per token at run time, so that a linear layer with weight $\bW$ and input $\bx$ computes $Q_4(\bW)\,Q_4(\bx)$.
Post-training quantizers choose the rounded weight $\hat\bW$ by minimizing the layer-wise reconstruction error $\tr\big((\hat\bW-\bW)\,\bH\,(\hat\bW-\bW)^\top\big)$, where $\bH = \mathbb{E}[\bx\bx^\top]$ collects the layer inputs on calibration data~\cite{frantar2023gptq}.
A WAM also caches keys and values: tokens that are computed once per control step, such as those of the observation or of the finished future, are read by attention at every denoising step that follows, and W4A4KV4 stores their keys and values at 4 bits as well~\cite{ashkboos2024quarot,liu2025spinquant}.

\noindent\textbf{Fresh and frozen errors.}
Rounding errors differ in how they recur over the denoising steps.
An error is \emph{fresh} if every step draws a new realization, as for the activations of tokens that are being denoised.
It is \emph{frozen} if one realization is reused at every step, as for a weight, a cached key or value, or a token that does not change across steps.

\noindent\textbf{Errors and the future share.}
We measure the action error $A = \|\hat\ba - \ba^\star\|^2$ on the executed part of the chunk, the future error $Z = \|\hat\bz - \bz^\star\|^2 / \|\bz^\star\|^2$ on the predicted frames, and the reader error
\begin{equation}
  K = \mathrm{KL}\big(\pi(R(\bz^\star)) \,\big\|\, \pi(R(\hat\bz))\big),
  \label{eq:reader-kl}
\end{equation}
where $R$ is the full-precision reader and $\pi$ maps each standardized action coordinate to a soft histogram; $K$ measures how much the future error alone changes the action.
A rounding error can change the action directly, or by degrading the predicted future on which the action is conditioned.
We measure the second path with the \emph{future share}
\begin{equation}
  \rho = 1 - \frac{\mathbb{E}\,\|\hat R(\bz^\star) - \ba^\star\|^2}{\mathbb{E}\,\|\hat R(\hat\bz) - \ba^\star\|^2},
  \label{eq:share}
\end{equation}
the fraction of the action error that disappears when the quantized reader is given the full-precision future.

\subsection{Observations}
\label{sec:observations}

We quantize Fast-WAM~\cite{yuan2026fastwam} to W4A4KV4 with rotation and second-order rounding~\cite{ashkboos2024quarot,frantar2023gptq}, protect the action pathway by keeping the activations of the action tokens at higher precision, and measure $A$, $Z$ and $\rho$ on LIBERO contexts.
We also inject errors of equal energy that are either frozen across the denoising steps or redrawn at every step.

\begin{figure}[t]
  \centering
  \includegraphics[width=0.92\linewidth]{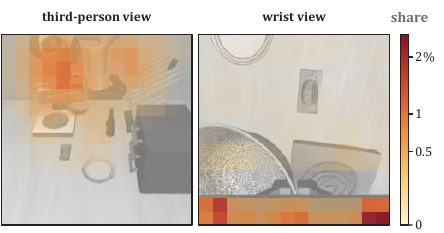}
  \caption{The reader depends on a narrow part of the future: share of the reader metric $\bM$ carried by each token of the first predicted frame of Fast-WAM, averaged over 80 contexts and shown over the frame of one of them.}
  \label{fig:obs:gaze}
\end{figure}

\noindent\textbf{Observation 1: Protecting the action is not enough.}
Sensitivities measured on the final action concentrate on the action pathway~\cite{akbari2026qwam}, which suggests protecting this pathway first.
Doing so removes only 16\% of the action error, and the future error does not change at all (\cref{tab:obs:swap}).
Restoring the predicted future to full precision removes 71.2\% of the remaining action error (\cref{fig:teaser}).
This implies that the quantization target must include the predicted future itself, rather than only the generated action.

\noindent\textbf{Observation 2: The future is damaged where it is written, and most by errors that are read many times.}
Raising the precision of the reader leaves the future error exactly unchanged, so the future can only be preserved in the computation that writes it.
Within this computation, rounding errors differ in how often they are read: the tokens of the observed frame keep the same rounding error at every denoising step, cached keys and values are written once and read at every step, and a weight is read by every token at every step.
An injected error of fixed energy damages the future 3.1$\times$, 2.2$\times$ and 2.1$\times$ more in the observation tokens, the observation cache and the text cache when it is frozen across the denoising steps than when it is redrawn at each step (\cref{tab:obs:reads}), because its effects on successive steps add up.
A sensitivity estimated one step at a time misses this accumulation: quantizing the ten video denoising steps together damages the part of the future that the reader depends on 2.50$\times$ more than the sum of the ten steps quantized one at a time (\cref{tab:obs:reads}, bottom).
This implies that an error should be weighted by how many times it is read, not only by the layer and step where it occurs.

\begin{table}[t]
  \centering
  \small
  \setlength{\tabcolsep}{5pt}
  \begin{tabular}{@{}lcc@{}}
    \toprule
    Model & $A$ ($\times 10^{-4}$) $\downarrow$ & $Z$ $\downarrow$ \\
    \midrule
    W4A4KV4                           & 3.53 & 0.0188 \\
    \quad + protected action pathway  & 2.96 & 0.0188 \\
    \quad + full-precision future     & 0.85 & 0 \\
    \bottomrule
  \end{tabular}
  \caption{Future swap on Fast-WAM. In the last row the quantized reader reads the future predicted by the BF16 model.}
  \label{tab:obs:swap}
\end{table}

\begin{table}[t]
  \centering
  \small
  \setlength{\tabcolsep}{5pt}
  \begin{tabular}{@{}lccc@{}}
    \toprule
    Injected error & Frozen & Redrawn & Ratio \\
    \midrule
    Observation tokens          & 11.56 & 3.68 & 3.1$\times$ \\
    Cache of the observed frame & 3.29  & 1.49 & 2.2$\times$ \\
    Cache of the text           & 0.48  & 0.23 & 2.1$\times$ \\
    \midrule
    Quantized video steps & Together & One at a time & Ratio \\
    \midrule
    All 10 denoising steps      & 35.0  & 14.0 & 2.50$\times$ \\
    \bottomrule
  \end{tabular}
  \caption{Future error weighted by the reader metric, $Z_{\bM}$ ($\times 10^{-3}$, \cref{eq:metric-form}).
  Top: an injected error of fixed energy, kept over the denoising steps (frozen) or redrawn at every read.
  Bottom: the 10 video denoising steps quantized together, against the sum of the 10 steps quantized one at a time.}
  \label{tab:obs:reads}
\end{table}

\noindent\textbf{Observation 3: The reader depends on a narrow part of the future, stable in location but not in direction.}
We linearize the reader at the full-precision future and compute its Gauss--Newton metric $\bM$ (\cref{sec:method:metric}).
In a 37{,}632-dimensional latent, $\bM$ has an effective rank of 41.5, and the first predicted frame carries 98\% of its trace.
Which locations of the latent the reader depends on is stable: the diagonal of $\bM$ correlates at 0.86 across disjoint halves of the calibration contexts, and it concentrates on a small region of the scene, the arm and the gripper (\cref{fig:obs:gaze}).
The directions are not: the leading 32 eigenvectors estimated on the two halves overlap by only 0.19.
This implies that protection should follow the tokens the reader depends on, rather than a fixed set of directions.

\section{Methodology}
\label{sec:method}

As illustrated in \cref{fig:method}, \method preserves the predicted future where it is written, stored and read.
A reader metric and a read count, both estimated offline, place the errors of activations and caches in one objective with a closed-form minimizer, and gaze-guided refinement gives the selected tokens a second 4-bit pass at run time.
\method involves no training.

\subsection{Reader metric}
\label{sec:method:metric}

The reader metric measures an error in the predicted future by how much it changes the action.
As Observation~3 shows (\cref{sec:observations}), the reader depends on a narrow part of the predicted future, whereas the uniform future error $Z$ weighs every latent dimension equally.
For a calibration context $c$, let $\bJ_c = \partial R(\bo_c, \ell_c, \bz)/\partial \bz$ at $\bz = \bz^\star_c$, and let $\bD_c$ be the diagonal curvature of the reader error $K$ (\cref{eq:reader-kl}) with respect to the reader output at $\ba^\star_c$.
We define the reader metric as the Gauss--Newton matrix of $K$ averaged over contexts, and the corresponding future error:
\begin{equation}
  \bM = \mathbb{E}_c\big[\bJ_c^\top \bD_c \bJ_c\big],
  \qquad
  Z_{\bM} = \frac{(\hat\bz - \bz^\star)^\top \bM\, (\hat\bz - \bz^\star)}{\|\bz^\star\|^2}
  \label{eq:metric}
\end{equation}
Here, $Z_{\bM}$ is the second-order change of the reader error caused by the future error $\hat\bz - \bz^\star$.
It reduces to the uniform future error $Z$ when $\bM = \mathbf{I}$.
A rounding error outside the reader changes the action only through the predicted future, so one metric prices the errors of every layer, token and denoising step that write the future, and estimating it involves the reader alone.

We never form $\bJ_c$ and estimate $\bM$ with Rademacher projections~\cite{hutchinson1989trace}: for a Rademacher vector $\boldsymbol{\epsilon}$, the probe $\mathbf{v} = \bJ_c^\top(\bD_c^{1/2}\boldsymbol{\epsilon})$ satisfies $\mathbb{E}[\mathbf{v}\mathbf{v}^\top] = \bJ_c^\top \bD_c \bJ_c$ and costs one reverse pass through the reader.
We normalize the estimate of each context to unit trace, average over contexts, keep the leading $r$ eigendirections together with the residual diagonal, which gives $\bM_r$, and mix in the identity:
\begin{equation}
  \bM = (1-\alpha)\,\frac{d}{\tr \bM_r}\,\bM_r + \alpha\,\mathbf{I},
  \label{eq:metric-form}
\end{equation}
where $d$ is the latent dimension and $\alpha \in (0, 1)$.
The diagonal is stable across contexts (Observation~3), the identity keeps every part of the future constrained, and $\bM$ is estimated once, on the full-precision model.

\subsection{Read count}
\label{sec:method:count}

The read count of an error source is the number of denoising steps over which one realization of its rounding error is read.
Observation~2 shows that this number changes the damage of an error, and that a sensitivity estimated one step at a time misses it.
Let $u$ index a source of rounding error: the activation of a token or a cached key or value.
Let $\be_{u,s}$ be the error that enters at denoising step $s$, and $\bJ_{u,s} = \partial\bz/\partial\be_{u,s}$ its effect on the predicted future.
To first order, $\hat\bz - \bz^\star = \sum_s \bJ_{u,s}\,\be_{u,s}$, so the future error in the reader metric is governed by the $T \times T$ matrix
\begin{equation}
  \bOmega_u(s, s') = \mathbb{E}\big[\tr\big(\bJ_{u,s}^\top\, \bM\, \bJ_{u,s'}\big)\big].
  \label{eq:kernel}
\end{equation}
For a fresh error with variance $\sigma_u^2$, the cross terms between steps vanish and the expected damage is $\sigma_u^2 \tr\bOmega_u$.
For a frozen error, $\be_{u,s} = \be_u$ at every step, every pair of steps contributes, and the damage is $\sigma_u^2\, \bone^\top \bOmega_u \bone$.
We define the read count as their ratio:
\begin{equation}
  n_u = \frac{\bone^\top \bOmega_u\, \bone}{\tr \bOmega_u} \in [0, T]
  \label{eq:count}
\end{equation}
A frozen error does the damage of a fresh error read $n_u$ times, and we set $n_u = 1$ for fresh sources.
Per-step sensitivities~\cite{akbari2026qwam,wang2026steerquant} keep only the diagonal of $\bOmega_u$, which corresponds to $n_u = 1$ for every source.

We measure $n_u$ by intervention rather than by differentiating through the sampler: we inject an error of fixed energy into $u$, once frozen across the denoising steps and once redrawn at every step, and take the ratio of the resulting $Z_{\bM}$, which holds at the error level of the 4-bit quantizer and not only to first order.
Sources of the same kind share one read count (each cache, the tokens that do not change across steps, and the tokens that are being denoised), so a few dozen forward passes suffice.

\begin{table*}[t]
  \centering
  \small
  \setlength{\tabcolsep}{3.5pt}
  \begin{tabular}{@{}llccccccccccccc@{}}
    \toprule
    & & & & & \multicolumn{5}{c}{LIBERO-Plus} & \multicolumn{5}{c}{LIBERO} \\
    \cmidrule(lr){6-10} \cmidrule(l){11-15}
    Model & Method & W & A & KV & Spatial & Object & Goal & Long & Avg. & Spatial & Object & Goal & Long & Avg. \\
    \midrule
    \multirow{6}{*}{Fast-WAM}
      & BF16                   & 16  & 16  & 16  & 81.4 & 82.9 & 67.6 & 78.6 & 77.6 & 98.6 & 99.5 & 97.6 & 96.2 & 98.0 \\
    \cmidrule(l){2-15}
      & QuaRot                 & 4.13 & 4  & 16  & 77.1 & 77.6 & 61.4 & 72.4 & 72.1 & 97.6 & 99.0 & 97.6 & 97.1 & 97.8 \\
      & SpinQuant              & 4.13 & 4  & 16  & 79.5 & 76.2 & 59.5 & 72.9 & 72.0 & 98.1 & 99.0 & 98.1 & 96.2 & 97.9 \\
      & ViDiT-Q                & 4.33 & 4  & 16  & 77.1 & 77.6 & 63.8 & 72.4 & 72.7 & 97.6 & 99.0 & 98.1 & 96.6 & 97.8 \\
      & SVDQuant               & 4.81 & 4  & 16  & 77.4 & 74.1 & 68.2 & 72.2 & \underline{73.0} & 97.1 & 100.0 & 98.1 & 97.1 & \underline{98.1} \\
      & \cellcolor{gray!12}\method & \cellcolor{gray!12}4.01 & \cellcolor{gray!12}4.62 & \cellcolor{gray!12}4.24 & \cellcolor{gray!12}82.4 & \cellcolor{gray!12}81.9 & \cellcolor{gray!12}64.3 & \cellcolor{gray!12}77.1 & \cellcolor{gray!12}\textbf{76.4} & \cellcolor{gray!12}98.1 & \cellcolor{gray!12}99.5 & \cellcolor{gray!12}98.1 & \cellcolor{gray!12}97.1 & \cellcolor{gray!12}\textbf{98.2} \\
    \midrule
    \multirow{6}{*}{Cosmos-Policy}
      & BF16                   & 16  & 16  & 16  & 84.3 & 89.5 & 73.3 & 86.7 & 83.5 & 98.6 & 100.0 & 98.6 & 98.1 & 98.8 \\
    \cmidrule(l){2-15}
      & QuaRot                 & 4.13 & 4  & 16  & 86.7 & 86.7 & 69.5 & 84.8 & 81.9 & 97.6 & 100.0 & 97.1 & 98.1 & 98.2 \\
      & SpinQuant              & 4.13 & 4  & 16  & 81.0 & 86.2 & 66.7 & 87.1 & 80.2 & 98.1 & 100.0 & 95.2 & 98.6 & 98.0 \\
      & ViDiT-Q                & 4.29 & 4  & 16  & 86.7 & 87.1 & 70.0 & 87.6 & \underline{82.9} & 97.6 & 99.5 & 97.1 & 99.5 & 98.5 \\
      & SVDQuant               & 4.89 & 4  & 16  & 84.3 & 88.1 & 72.9 & 86.2 & \underline{82.9} & 99.0 & 99.5 & 97.1 & 99.5 & \textbf{98.8} \\
      & \cellcolor{gray!12}\method & \cellcolor{gray!12}4.01 & \cellcolor{gray!12}4.58 & \cellcolor{gray!12}4.70 & \cellcolor{gray!12}86.7 & \cellcolor{gray!12}89.5 & \cellcolor{gray!12}68.6 & \cellcolor{gray!12}88.6 & \cellcolor{gray!12}\textbf{83.3} & \cellcolor{gray!12}98.6 & \cellcolor{gray!12}100.0 & \cellcolor{gray!12}98.6 & \cellcolor{gray!12}97.6 & \cellcolor{gray!12}\underline{98.7} \\
    \bottomrule
  \end{tabular}
  \caption{Closed-loop success rates (\%) on LIBERO-Plus and LIBERO.
  W, A and KV give the average bits of the weights (including their scales and any high-precision component), of the activations, and of the cached keys and values: the baselines keep their caches in BF16, and \method also quantizes them to 4 bits.
  \textbf{Bold}: best quantized result for each model; \underline{underline}: second best.
  }
  \label{tab:main}
\end{table*}

\subsection{Joint allocation}
\label{sec:method:alloc}

Joint allocation decides, under one budget, which tokens receive more precision in the denoising loop and in the cache.
Observation~1 shows that the target must include the predicted future, so \method keeps the quantized model close to the full-precision one in
\begin{equation}
  \mathcal{L} = \mathbb{E}_c\big[\,A + \lambda_{\bM} Z_{\bM} + \beta K\,\big],
  \label{eq:objective}
\end{equation}
where the three terms cover the action, the future as the reader sees it, and the action read from the future.
We fix $\lambda_{\bM}$ and $\beta$ once from the initial magnitudes of the three terms.
To second order in the rounding errors, $\mathcal{L}$ separates over their sources:
\begin{equation}
  \mathcal{L} \approx C_{\bW} + \sum_{u} n_u\, \omega_u\, \sigma_u^2(r_u),
  \label{eq:separable}
\end{equation}
where $C_{\bW}$ collects the errors of the 4-bit weights, which refinement does not change.
The sum runs over the units of activations and caches, a token at a denoising step or a cached token, with sensitivity $\omega_u = \tr\bOmega_u$, error variance $\sigma_u^2$, and $r_u \in \{0, 1\}$ indicating whether unit $u$ is refined (\cref{sec:method:refine}).

Refining unit $u$ divides $\sigma_u^2$ by at least 196 (\cref{eq:refine}) at a cost $c_u$: one more row of a 4-bit matrix product for an activation, one more 4-bit code for a cached entry.
Minimizing \cref{eq:separable} under a budget $\sum_u c_u r_u \le B$ is solved by a threshold on the damage removed per unit cost:
\begin{equation}
  r_u = 1 \iff \frac{n_u\, \omega_u\, \sigma_u^2}{c_u} \ge \mu
  \label{eq:threshold}
\end{equation}
where $\mu$ is set by $B$.
One threshold serves activations and caches alike, so precision is traded between the denoising loop and the cache in a single currency.

For the token at position $j$ and denoising step $s$, we factor the sensitivity as $\omega_{s,j} = \phi_s\, g_j$, where the step profile $\phi_s$ is measured by refining all tokens at step $s$ alone and $g_j$ is the gaze of \cref{sec:method:refine}.
Cached tokens and tokens that do not change across steps have large read counts and are written once, so they are refined first; the tokens that are being denoised follow, at the denoising steps and token positions where the product $\phi_s\, g_j$ is largest.

\subsection{Gaze-guided refinement}
\label{sec:method:refine}

Gaze-guided refinement gives the selected tokens a second 4-bit pass and follows the attention of the reader to decide which tokens these are.
Observation~3 shows that the locations the reader depends on are stable while the directions are not, so we protect tokens rather than directions, and we do so without leaving 4-bit arithmetic.

For the activation $\bx$ of a selected token, we quantize the residual of the first pass a second time:
\begin{equation}
  \begin{gathered}
    \mathbf{q}_1 = \operatorname{round}\!\Big(\frac{\bx}{\varsigma_1}\Big), \qquad
    \mathbf{q}_2 = \operatorname{round}\!\Big(\frac{\bx - \varsigma_1 \mathbf{q}_1}{\varsigma_2}\Big), \\[4pt]
    \hat\bx = \varsigma_1 \mathbf{q}_1 + \varsigma_2 \mathbf{q}_2
  \end{gathered}
  \label{eq:refine}
\end{equation}
with $\varsigma_1$ the scale of \cref{sec:prelim} and $\varsigma_2 = \max|\bx - \varsigma_1 \mathbf{q}_1| / 7 \le \varsigma_1 / 14$.
The rounding error is bounded by $\varsigma_1/28$ instead of $\varsigma_1/2$, which lowers its power by a factor of at least 196.
The layer output becomes $\varsigma_1\, Q_4(\bW)\,\mathbf{q}_1 + \varsigma_2\, Q_4(\bW)\,\mathbf{q}_2$: two 4-bit matrix products, the second over the selected rows only.
For a selected token in a cache, the second code $\mathbf{q}_2$ is stored next to the first.
No weight is stored at higher precision, and the quantized model has no branch outside the 4-bit path.

The \emph{gaze} $g_j$ of token $j$ is its share of the reader's sensitivity.
Offline, it is the trace of the block of $\bM$ that belongs to token $j$.
At run time the reader provides it for free: the attention that the action tokens pay to token $j$, summed over layers, heads and action denoising steps, is a by-product of generating the action.
Because consecutive control steps see nearly the same scene, the gaze of one control step selects the tokens that are refined at the next, and the first control step of an episode uses the offline gaze.
Observation tokens inherit the gaze through the attention of the future tokens, and when the future and the action are denoised jointly, the gaze of one denoising step selects the tokens of the next.
The selection therefore follows the reader in the current scene, and the quantities fixed offline ($\bM$, $n_u$, $\phi_s$ and $\mu$) do not depend on the task, so \method needs no recalibration for a new task.

\section{Experiments}
\label{sec:exp}

\subsection{Experimental setup}
\label{sec:exp:setup}

\noindent\textbf{Models and benchmarks.}
We quantize Fast-WAM~\cite{yuan2026fastwam} (6B parameters, of which 1B form the action expert) and Cosmos-Policy~\cite{kim2026cosmospolicy} (2B parameters).
Fast-WAM first denoises the future latent and then denoises an action chunk that attends to it through a key--value cache; the tokens of its observed frame do not depend on the denoising step and are computed once.
Cosmos-Policy denoises the future and the action jointly, so its reader starts from the future state at an intermediate denoising step and runs the remaining steps to the action, and its keys and values are quantized at every step.
We evaluate in simulation on LIBERO-Plus~\cite{fei2026liberoplus}, which perturbs the four LIBERO suites~\cite{liu2023libero} along seven dimensions, on the original LIBERO suites, and on a real robot arm (\cref{sec:exp:real}).
LIBERO-Plus is our main benchmark, since both BF16 models already succeed in about 98\% of the episodes of the original LIBERO suites.

\noindent\textbf{Baselines and quantization settings.}
We compare with the general-purpose quantizers QuaRot~\cite{ashkboos2024quarot} and SpinQuant~\cite{liu2025spinquant}, and the diffusion quantizers ViDiT-Q~\cite{zhao2025viditq} and SVDQuant~\cite{li2025svdquant}; the WAM quantizers~\cite{zhou2026quantwams,akbari2026qwam,wang2026steerquant} have no public implementation.
All baselines quantize weights and activations to 4 bits and keep the cached keys and values in BF16, whereas \method also quantizes the caches to 4 bits.
Since adaptive layer normalization prevents a rotation of the residual stream, QuaRot and SpinQuant rotate the input of every linear layer by block rotations of size 128; ViDiT-Q and SVDQuant follow their published recipes, the latter with a 16-bit low-rank branch of rank 32.
\method uses 4-bit weights rounded by GPTQ~\cite{frantar2023gptq} in randomized Hadamard coordinates, 4-bit activations with per-token scales, and 4-bit keys and values with one scale per token and head.
As is common, all methods keep the input and output interface layers (0.01\% of the weights) in BF16 and store the timestep-conditioning layers as exact per-step tables.
ViDiT-Q, SVDQuant and \method calibrate on the same 80 LIBERO contexts, and the budget of \method refines 15.4\% of the activation values of the denoising loops and 6.0\% of the cached keys and values.

\subsection{Main results}
\label{sec:exp:main}

\noindent\textbf{Results on Fast-WAM.}
On LIBERO-Plus (\cref{tab:main}), \method reaches 76.4\% average success, within 1.2 points of BF16 (77.6\%) and 3.4 points above SVDQuant, the strongest baseline, although the baselines keep their caches in BF16.
The baselines lose 4.6--5.6 points relative to BF16.
The gains of \method concentrate on the camera-viewpoint and sensor-noise perturbations, where the baselines lose the most.
On LIBERO, where BF16 reaches 98.0\%, the baselines stay within 0.2 points of it, and \method reaches 98.2\%, above every baseline and level with BF16.

\noindent\textbf{Results on Cosmos-Policy.}
On LIBERO-Plus, \method reaches 83.3\% average success, level with BF16 (83.5\%) and 0.4 points above ViDiT-Q and SVDQuant, the strongest baselines.
The baselines lose 0.6--3.3 points, mostly on the Goal suite.
On LIBERO, where BF16 reaches 98.8\%, the baselines lose up to 0.8 points, SpinQuant again the most, and \method reaches 98.7\%, level with BF16 and with SVDQuant, the strongest baseline.
On the Goal suite, where every baseline loses 1.5 points or more, \method keeps the 98.6\% of BF16.

\begin{figure}[t]
  \centering
  \includegraphics[width=\linewidth]{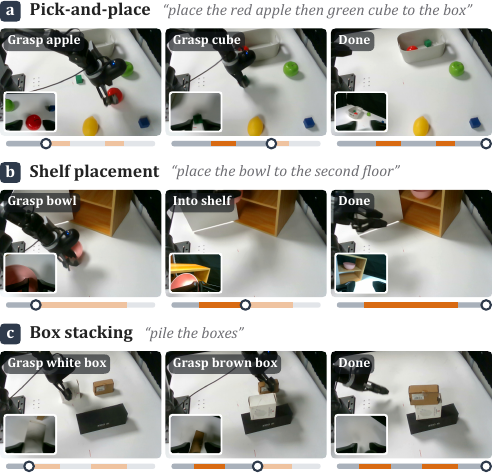}
  \caption{Real-world tasks on the AgileX PiPER arm: three key frames of one demonstration each, with the wrist view inset.
  The bar under a frame spans the demonstration (orange: gripper closed; dot: time of the frame).}
  \label{fig:real}
\end{figure}

\subsection{Real-world deployment}
\label{sec:exp:real}

\noindent\textbf{Setup.}
We deploy Fast-WAM on an AgileX PiPER arm (6 degrees of freedom and a parallel gripper).
We fine-tune it on teleoperation data collected with LeRobot~\cite{cadene2024lerobot} from a third-person and a wrist camera, with 75 to 80 demonstrations per task.
We quantize the fine-tuned BF16 model to W4A4KV4 with calibration contexts drawn from the demonstrations and re-estimate the reader metric on it.
We compare BF16, QuaRot with the settings of \cref{tab:main} and \method with 25 trials per task and method on the same set of marked initial object placements on the table.

\noindent\textbf{Tasks.}
\cref{fig:real} shows the three tasks: \emph{two-object pick-and-place}, which puts a red apple and then a green cube into a box among three distractors; \emph{shelf placement}, which places a bowl on the second floor of a shelf; and \emph{box stacking}, which stacks a white box on a black base and then a brown box on the white one.

\begin{table}[t]
  \centering
  \small
  \setlength{\tabcolsep}{4pt}
  \begin{tabular}{@{}lccccc@{}}
    \toprule
    & \multicolumn{2}{c}{Pick-and-place} & & & \\
    \cmidrule(lr){2-3}
    Method & First & Both & Shelf & Stacking & Avg. \\
    \midrule
    BF16    & 19/25 & 2/25 & 15/25 & 16/25 & 52.0\% \\
    QuaRot  & 4/25 & 2/25 & 3/25 & 8/25 & 17.0\% \\
    \method & 15/25 & 4/25 & 13/25 & 15/25 & 47.0\% \\
    \bottomrule
  \end{tabular}
  \caption{Real-world results on the AgileX PiPER arm (successes / trials).
  For pick-and-place, \emph{First} counts trials in which the apple reaches the box and \emph{Both} trials in which both objects do; the average is over the four columns.}
  \label{tab:real}
\end{table}

\noindent\textbf{Results.}
\cref{tab:real} summarizes the real-world results.
\method reaches 47.0\% average success, against 17.0\% for QuaRot and 52.0\% for BF16.
On pick-and-place, it places the apple in 15 of 25 trials and both objects in 4, against 4 and 2 for QuaRot.
On shelf placement and box stacking, it stays within two trials of BF16 (13 and 15 against 15 and 16), while QuaRot succeeds in 3 and 8.
The closed-loop gains of \method under 4-bit quantization thus carry over from simulation to a physical robot arm.

\subsection{Ablation study}
\label{sec:exp:ablation}

\begin{table}[t]
  \centering
  \small
  \setlength{\tabcolsep}{2.6pt}
  \begin{tabular}{@{}lccccc@{}}
    \toprule
    & \multicolumn{2}{c}{Bits} & \multicolumn{2}{c}{Error $\downarrow$} & Succ. \\
    \cmidrule(lr){2-3} \cmidrule(lr){4-5}
    Variant & A & KV & Action & Reader & (\%) $\uparrow$ \\
    \midrule
    \multicolumn{6}{@{}l}{\emph{Refined tokens}} \\
    \hspace{0.5em}None                               & 4    & 4    & 4.70 & 1.75 & 71.2 \\
    \hspace{0.5em}Action tokens (BF16)               & 4.71 & 4    & 4.18 & 1.75 & 71.9 \\
    \hspace{0.5em}Observation tokens                 & 4.55  & 4.12  & 2.62 & 1.08 & 74.3 \\
    \midrule
    \multicolumn{6}{@{}l}{\emph{Selection criterion, same budget}} \\
    \hspace{0.5em}Random                             & 4.61 & 4.33 & 3.19 & 1.23 & 72.5 \\
    \hspace{0.5em}Per-step sensitivity               & 4.62  & 4.24  & 2.71 & 1.52 & 72.9 \\
    \hspace{0.5em}Offline gaze                       & 4.62 & 4.24 & 2.28 & 0.96 & 75.2 \\
    \midrule
    \multicolumn{6}{@{}l}{\emph{Component removed, same budget}} \\
    \hspace{0.5em}Reader metric ($\bM = \mathbf{I}$) & 4.62  & 4.24  & 2.74 & 1.31 & 73.6 \\
    \hspace{0.5em}Read count ($n_u = 1$)             & 4.62  & 4.24  & 2.55 & 1.17 & 74.0 \\
    \midrule
    \multicolumn{6}{@{}l}{\emph{Cache precision}} \\
    \hspace{0.5em}BF16 caches                        & 4.62 & 16   & 2.02 & 0.78 & 76.8 \\
    \midrule
    \cellcolor{gray!12}\method & \cellcolor{gray!12}4.62 & \cellcolor{gray!12}4.24 & \cellcolor{gray!12}2.10 & \cellcolor{gray!12}0.82 & \cellcolor{gray!12}76.4 \\
    \bottomrule
  \end{tabular}
  \caption{Ablations on Fast-WAM.
  A and KV: average bits of the activations and of the cached keys and values; \emph{Action} and \emph{Reader}: action error $A$ ($\times 10^{-4}$) and reader error $K$ ($\times 10^{-2}$) on held-out LIBERO contexts; \emph{Succ.}: closed-loop success rate on LIBERO-Plus.}
  \label{tab:ablation}
\end{table}

\cref{tab:ablation} isolates each component of \method.

\noindent\textbf{Exact actions are not enough.}
Plain W4A4KV4 loses 6.4 points to BF16.
Keeping the action tokens in BF16 recovers only 0.7 of them although its action tokens are exact: the action error falls by only 11\%, and the reader error does not change at all, since the exact action tokens read the same degraded future.
This confirms Observation~1 in closed loop.
Refining only the observation tokens, which are written once and read at every denoising step, recovers 3.1 points, and the future tokens and cached entries selected by the gaze bring \method to 76.4\%, with both errors below half of those of plain W4A4KV4.

\noindent\textbf{The reader decides which tokens matter.}
With the pipeline, the budget and the refinement fixed, random tokens leave 1.5$\times$ the action error and the reader error of \method and reach 72.5\%.
A per-step sensitivity of the final action, the kind of criterion that existing WAM quantizers use~\cite{akbari2026qwam,wang2026steerquant}, depends on the action alone and reaches 72.9\%: it spends the budget on the action pathway and lowers the reader error by only 13\%.
The offline gaze reaches 75.2\%, and tracking the gaze at run time lowers the reader error by a further 15\% and adds 1.2 points, mostly on the camera-viewpoint perturbations.
Criteria computed from the video model itself, its activation energy~\cite{federici2026stamp} and its self-attention~\cite{shen2024squat,jang2026semanticdialect}, fall behind the gaze as well.

\noindent\textbf{The reader metric and the read count both matter.}
Replacing the reader metric by the identity, so that every part of the predicted future counts equally, lowers the success rate from 76.4\% to 73.6\%.
Setting $n_u = 1$ for every source moves the budget from the tokens that are read at every denoising step to the tokens that are being denoised, and lowers it to 74.0\%.

\noindent\textbf{Cache precision.}
Keeping the caches of \method in BF16 adds only 0.4 points, so its 4-bit caches cost little.

\begin{table}[t]
  \centering
  \small
  \setlength{\tabcolsep}{3pt}
  \begin{tabular}{@{}lcccc@{}}
    \toprule
    Method & Weights & Caches & Peak mem. & Latency \\
           & (GB)    & (MB)   & (GB)      & (ms) \\
    \midrule
    BF16                       & 12.0 & 204 & 14.0 & 1475 \\
    \method                    & 3.2  & 56  & 6.0  & 888 \\
    \bottomrule
  \end{tabular}
  \caption{Cost of one Fast-WAM policy call on an NVIDIA A100.
  Weights cover the whole denoiser, caches are the keys and values of one call, peak memory covers the whole policy with the text embedding of the instruction computed once, and latency is measured end to end from observation to action chunk.}
  \label{tab:efficiency}
\end{table}

\subsection{Efficiency}
\label{sec:exp:efficiency}

\cref{tab:efficiency} reports the cost of \method on Fast-WAM on an NVIDIA A100, with its weights stored and multiplied as 4-bit codes.
With 4.01 bits per weight and no high-precision branch, \method compresses the denoiser from 12.0\,GB to 3.2\,GB (3.8$\times$) and reduces the peak GPU memory of the whole policy from 14.0\,GB to 6.0\,GB (2.3$\times$).
The cached keys and values of one call shrink from 204\,MB to 56\,MB (3.6$\times$).
One policy call takes 888\,ms from observation to action chunk, 1.66$\times$ faster than the BF16 model (1475\,ms): the runtime multiplies the packed 4-bit codes directly on the integer tensor cores of the GPU, and the second pass of a refined token is the same integer product on additional rows.
Calibration takes 1.2 GPU-hours with a peak memory of 27\,GB, since it only runs forward passes of the quantized model on calibration contexts.

\section{Conclusion}
\label{sec:conclusion}

We presented \method, a latent-aware quantization framework for 4-bit world action models.
Once the action pathway of a 4-bit WAM is protected, its remaining action error enters through a degraded predicted future.
\method measures future error as the reader sees it, weights each rounding error by how many times it is read, allocates precision jointly over activations and caches, and gives the tokens the reader attends to a second 4-bit pass.
With every weight at 4 bits and no recalibration per task, it recovers closed-loop success across two WAMs in simulation and on a real robot arm.
When compressing a WAM, the predicted future is not an intermediate quantity that can be sacrificed.

{
    \small
    \bibliographystyle{ieeenat_fullname}
    \bibliography{main}
}

\end{document}